\documentclass{article}
\usepackage{spconf,amsmath,amssymb,graphicx}

\title{Localizing Catastrophic Forgetting in Neural Networks}
\name{Felix Wiewel and Bin Yang}
\address{Institute of Signal Processing and System Theory, University of Stuttgart, Germany}
\begin{document}
%
\maketitle
\begin{abstract}
\textbf{Artificial neural networks (ANNs) suffer from catastrophic forgetting when trained on a sequence of tasks. While this phenomenon was studied in the past, there is only very limited recent research on this phenomenon. We propose a method for determining the contribution of individual parameters in an ANN to catastrophic forgetting. The method is used to analyze an ANNs response to three different continual learning scenarios.}
\end{abstract}
\section{Introduction}
\label{sec:intro}

Artificial neural networks suffer from a phenomenon called catastrophic forgetting, which is characterized by a rapid decrease in performance on a learned task when trained on a new task \cite{ratcliff1990connectionist, mccloskey1989catastrophic}. For example an ANN trained on machine translation between English and German will essentially "forget" everything it has learned when it is trained on translating between German and French. This is in contrast to human learning, where a human typically will remember at least something he or she has learned on a past task. Solving this problem of catastrophic forgetting, i.e. enabling continual learning of ANNs, is of great interest, because it can enable the accumulation of knowledge over possibly long periods of time without requiring training examples from all but the most recent task. This comes with a number of benefits compared with the current standard of jointly training an ANN on all tasks simultaneously. First: Since training would not require examples from all previously learned tasks, data for a task that has already been learned is no longer needed and can be discarded, reducing the memory required for training. Second: After a ANN was trained to solve some tasks, it is not static but can be adjusted to solve newly and potentially unforeseen tasks. Third: The overall time required for training an ANN on a sequence of tasks could be reduced, since it only needs to be trained on the new task without retraining with the data of previously learned tasks.\\\\
While catastrophic forgetting is known in the literature since 1989 and has been studied quite intensively in the past \cite{robins1996consolidation, yamaguchi2004reassessment, french1999catastrophic, hetherington1993catastrophic}, interest in this phenomenon has decayed over the years. Only recently there has been a renewed interest in solving this problem. Many new methods for overcoming catastrophic forgetting like \textit{Elastic Weight Consolidation} (EWC), \textit{Synaptic Intelligence} (SI), \textit{Deep Generative Replay} (DGR), \textit{Variational Continual Learning} (VCL) and more have been proposed \cite{kirkpatrick2017overcoming, zenke2017continual, shin2017continual, v.2018variational}. Although these works propose new ways of mitigating catastrophic forgetting there is only a very limited research on the phenomenon itself. One example of this is an empirical study on catastrophic forgetting by Goodfellow et al. \cite{goodfellow2013empirical}. In their work the authors compare different activation functions and their effect on mitigating catastrophic forgetting. The choice of activation function is a vital part of designing a neural architecture and its resilience to catastrophic forgetting is import, but it does not give an insight into the internal mechanisms of an ANN.\\\\
In this paper, we study catastrophic forgetting in ANNs by quantifying which part of the network contributed with what extend to forget a previously learned task. While catastrophic forgetting in previous works is measured as a scalar value, e.g. the increase of loss or decrease of accuracy on a previously learned task, we propose a method to quantify catastrophic forgetting separately for every parameter in an ANN. This not only allows for a coarse analysis, i.e. if a ANN experiences catastrophic forgetting or not, but it also localizes which part of a neural architecture contributes to which extend of forgetting a previously learned task. We think that a deeper understanding of catastrophic forgetting in ANNs, enabled through this work, can lead to better methods for overcoming it.

\section{Methods}
\label{sec:metho}

In this section we describe the notation, define different scenarios and methods used in this work.

\subsection{Notation}
\label{ssec:notat}

In order to clearly define a sequence of tasks on which a ANN is trained, we borrow the notation used in the remainder of this paper from the closely related field of transfer learning with a slight modification\footnote{We use the term "assignment" for what is referred to as a "task" in transfer learning.} in order to avoid confusion \cite{weiss2016survey}. We start by defining a domain $\mathcal{D}$ which consists of two parts, a feature space $\mathcal{X}$ and a marginal data generating distribution $P(\mathbf{X})$, where $\mathbf{X}=\lbrace\mathbf{x}_{1},\ldots,\mathbf{x}_{N}\rbrace\in\mathcal{X}$ is a set of training examples. In image classification, the feature space is given by $\mathcal{X}=\lbrace 0,1,\ldots,255\rbrace^{N\times C}$, where $N$ and $C$ are the number of pixels and channels an image contains. An assignment $\mathcal{A}$ for a given domain $\mathcal{D}$ is again defined by two parts, a label space $\mathcal{Y}$ and a function $f:\mathcal{X}\rightarrow\mathcal{Y}$, which represents the mapping from feature to label space. The function $f$ is learned from pairs $\lbrace x_{i},y_{i}\rbrace$, where $x_{i}\in\mathcal{X}$ is a training example and $y_{i}\in\mathcal{Y}$ is the corresponding label. In image classification, this function maps an image to its label. With this notation we can define the phenomenon of catastrophic forgetting more precisely as a rapid decrease in performance of an ANN on Task $A$, defined by $\mathcal{D}_{A}$ and $\mathcal{A}_{A}$, as it is trained on Task $B$, defined by $\mathcal{D}_{B}$ and $\mathcal{A}_{B}$, if $\mathcal{D}_{B}\neq\mathcal{D}_{A}$ and/or $\mathcal{A}_{B}\neq\mathcal{A}_{A}$.

\subsection{Continual Learning Scenarios}
\label{ssec:CLsce}

Although the recent work on continual learning shares the same goal of mitigating catastrophic forgetting, different experimental setups are used to evaluate the proposed methods. These differ significantly and pose different challenges to the algorithms and methods which are evaluated on them. In order to make the research in this area more comparable, three different scenarios of continual learning were recently proposed \cite{Hsu2018ReevaluatingCL, Ven2018GenerativeRW}. These categories are defined in the following subsections.

\subsubsection{Incremental Domain Learning}
\label{sssec:IncDo}

Incremental domain learning (IDL) is characterized by a change in at least one part of the domain $\mathcal{D}$, either the feature space $\mathcal{X}$, the data generating distribution $P(\mathbf{X})$ or both change. This scenario is similar but not identical to domain adaptation in the field of transfer learning \cite{weiss2016survey}. The difference between domain adaptation and IDL is given by the fact, that in domain adaption one is only interested in transferring an ANNs knowledge from domain $\mathcal{D}_{A}$ to $\mathcal{D}_{B}$. After this transfer the ANNs performance on Domain $\mathcal{D}_{A}$ is typically irrelevant. In IDL this is not the case. Here one is interested learning to solve a task on domain $\mathcal{D}_{A}$ and on $\mathcal{D}_{B}$ without catastrophic forgetting and possibly a transfer of knowledge between the two domains. Another important property of IDL is that the assignment remains unchanged, i.e. the label space $\mathcal{Y}$ and the function $f$ remain unchanged. This can be represented more formally with $f(\mathbf{x}_{i})=f(\hat{\mathbf{x}}_{i})=\mathbf{y}_{i}$, where $\hat{\mathbf{x}}_{i}$ is the representation of $\mathbf{x}_{i}$ in a different domain. In practice this means, that we can share the same output layer of an ANN over different domains. A widely used example for this scenario is permutation MNIST \cite{kirkpatrick2017overcoming, zenke2017continual, shin2017continual, v.2018variational}. In order to generate different domains, random pixel permutations of images in the MNIST are used, where one realization of a permutation is applied to all images. Although this does not change the feature space $\mathcal{X}$, it changes the data generating distribution $P(\mathbf{X})$ and hence the domain $\mathcal{D}$. The random permutations used in this example generate uncorrelated domains, which is not very realistic and has caused some criticism \cite{zenke2017continual, Hsu2018ReevaluatingCL}. A very simple example of IDL with highly correlated, and therefore more realistic, domains based on the MNIST data set can be generated by just inverting the pixel intensities.

\subsubsection{Incremental Class Learning}
\label{sssec:IncCl}

In incremental class learning (ICL) each task adds one or possibly more new classes to classify by an ANN. In each task the ANN is presented with a data set containing only examples of at least one new class to learn. This means that not only the domain $\mathcal{D}$ but also the assignment $\mathcal{A}$ changes between tasks. Formally, the feature space $\mathcal{X}$ and/or the data generating distribution $P(\mathbf{X})$ and the function $f$ change. The label space $\mathcal{Y}$ remains unchanged and therefore the output layer of the ANN can also be shared between tasks. A widely used example for this scenario is split MNIST \cite{shin2017continual}. In order to generate a sequence of tasks, the MNIST data set is split in such a way that each split contains all the examples from at least one class. A typical way to split the MNIST data set is to separate it into $5$ disjoint sets each containing two classes, e.g. $(0,1)$, $(2,3)$, $(4,5)$, $(6,7)$ and $(8,9)$. In practice this means that in the first task only examples of the classes $0$ and $1$ are used to train an ANN while its output layer contains $10$ neurons and is therefore able to distinct between $10$ different classes.

\subsubsection{Incremental Task Learning}
\label{sssec:IncTa}

The last scenario, incremental task learning (ITL), also allows for changes in both the domain $\mathcal{D}$ and the assignment $\mathcal{A}$ between tasks. In contrast to ICL the label space $\mathcal{Y}$ also changes in ITL, i.e. a ANN can first learn a classification and then a regression task. Since these tasks typically require different activation functions for the last layer of an ANN, a new output layer is used for every task. Such an ANN is known as a multi-headed ANN in the continual learning literature \cite{v.2018variational, Hsu2018ReevaluatingCL}. This implies that during inference the identity of a task, which needs to be solved by an ANN, is known in order to select the corresponding head. Requiring such a prior knowledge about the task identity is in stark contrast to ICL, where the ANN not only solves the task at hand but also infers which task is to be solved. A typical example for a sequence of tasks for ITL can be generate from split MNIST or a set of different data sets, where each split or data set has to be learned in each task, by using a different output layer for each task.\\\\

\noindent These three scenarios for continual learning differ not only in their setup but also in the challenges they pose to an ANN. While ICL requires the model not only to solve a task but also to recognize which task, from all the previously learned tasks, it has to solve. IDL on the other hand requires an ANN to solve the same task across different domains. The last scenario ITL just requires a model to solve individual tasks while sharing only a subset of the network across them.

\subsection{Proposed Method}
\label{ssec:PropM}

\begin{figure}
	\centering
	\includegraphics[width=1.0\linewidth]{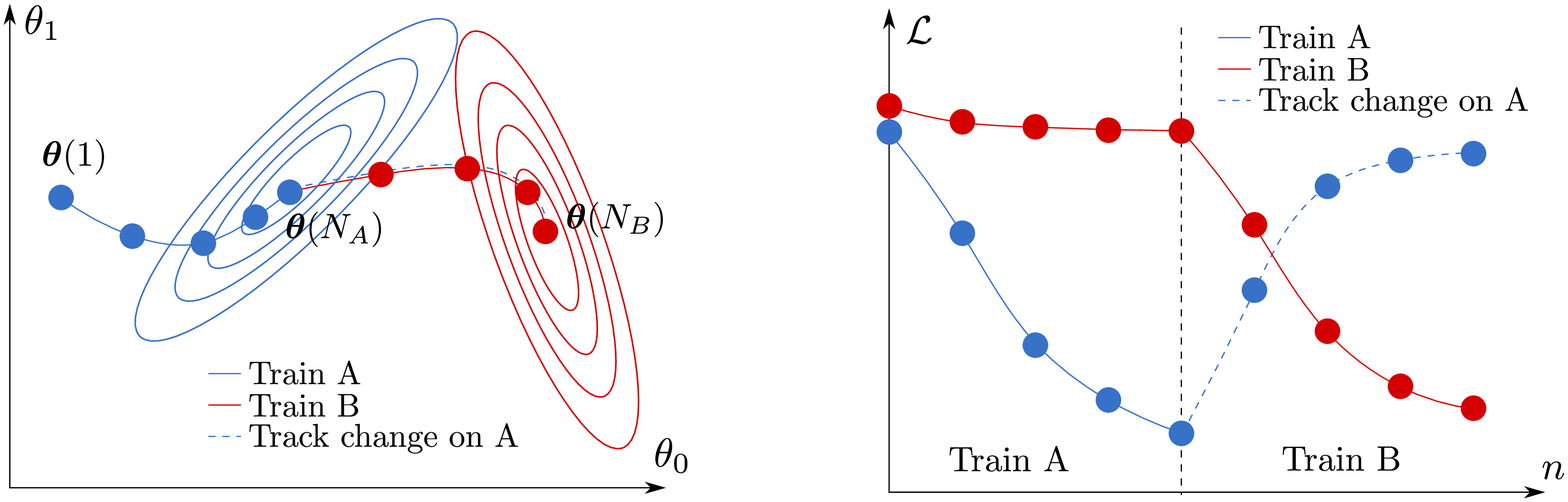}
	\caption[Proposed method]{Proposed method for localizing catastrophic forgetting}
	\label{fig:method}
\end{figure}

In order to quantify catastrophic forgetting we need a measure for the performance of an ANN on a given task. Since ANNs are typically trained using stochastic gradient descent, or some variant of it, to minimize a defined loss function $\mathcal{L}(\boldsymbol{\theta}, \mathcal{D}, \mathcal{A})$ with respect to its parameters $\boldsymbol{\theta}$, it is a natural choice to use this loss function for quantifying catastrophic forgetting. To simplify the notation and reduce clutter we will use $\mathcal{L}_{A}(\boldsymbol{\theta})$ for a loss function $\mathcal{L}(\boldsymbol{\theta}, \mathcal{D}_{A}, \mathcal{A}_{A})$, which is defined on domain $\mathcal{D}_{A}$ with assignment $\mathcal{A}_{A}$.\\\\
Similar to the inspiring work of Zenke et al. \cite{zenke2017continual} we interpret the training process of an ANN as a trajectory in parameter space defined by $\boldsymbol{\theta}(n)$, where $n\in\mathbb{N}^{+}$ is the current training step. Moving the parameters $\boldsymbol{\theta}$ along this trajectory causes a change in the loss function $\Delta\mathcal{L}$. If we compute the gradient $\boldsymbol{\nabla}_{\boldsymbol{\theta}}\mathcal{L}$ at each point in parameter space, we can compute $\Delta\mathcal{L}$ either through the difference in loss between the start- and endpoint or through the path integral of $\boldsymbol{\nabla}_{\boldsymbol{\theta}}\mathcal{L}$ along the trajectory as
\begin{align}
\label{eq:deltaL}
	\Delta\mathcal{L}=\mathcal{L}(\boldsymbol{\theta}(N))-\mathcal{L}(\boldsymbol{\theta}(1))=\int_{\mathcal{C}}\boldsymbol{\nabla}_{\boldsymbol{\theta}}\mathcal{L}(\boldsymbol{\theta})\mathrm{d}\boldsymbol{\theta},
\end{align}
where $N$ is the number of training steps and $\mathcal{C}$ is the trajectory of $\boldsymbol{\theta}$ through parameter space during training. This equivalence holds, since the gradient vector field is a conservative field. Although both methods for computing this change yield the same result, they differ in their complexity and the insight they can provide. While evaluating $\Delta\mathcal{L}$ via the difference in loss at the start- and endpoint is fast and simple, it can only provide information about the ANN as a whole. Using the path integral is computationally more expensive and therefore slower, but it enables us to determine the contribution of individual parameters. In order to calculate a parameter specific contribution to the change in loss, we decompose the path integral and approximate it with a sum as
\begin{align}
	\Delta\mathcal{L}&=\int_{\mathcal{C}}\sum_{i}\boldsymbol{\nabla}_{\theta_{i}}\mathcal{L}(\boldsymbol{\theta})\mathrm{d}\theta_{i}\nonumber\\&\approx\sum_{j=1}^{N}\sum_{i}\boldsymbol{\nabla}_{\theta_{i}}\mathcal{L}(\boldsymbol{\theta}_{j})\Delta\theta_{ij}=\sum_{i}\Delta\mathcal{L}_{i},
\end{align}
where $\Delta\theta_{ij}$ is a small change in the $i$th parameter at training step $j$. With this approximation we can determine the individual contribution of the $i$th parameter $\Delta\mathcal{L}_{i}$ to the overall change in loss $\Delta\mathcal{L}$. In order to check if the approximation is accurate, we can use equation \ref{eq:deltaL} to compute $\Delta\mathcal{L}$ exactly and compare it with our approximation. In general the accuracy will depend on the curvature of the loss surface and the step size used for parameter updates. If there is an unacceptable difference between the exact change and the proposed approximation, one can improve the accuracy by inserting some intermediate steps for evaluation of the path integral between two parameter updates.\\\\
Catastrophic forgetting occurs when we transition from training and ANN on minimizing $\mathcal{L}_{A}(\boldsymbol{\theta})$ to it being trained to minimize $\mathcal{L}_{B}(\boldsymbol{\theta})$ and is characterized by a rapid increase of the former right after the transition. This period of rapid change is of particular interest, since it represents the period over which the ANN forgets a previously learned task. Determining the contributions of individual parameters is therefore most useful right after the transition and has to be done for the loss function $\mathcal{L}_{A}(\boldsymbol{\theta})$. This process is illustrated in Fig. \ref{fig:method}.\\\\
Although we limit our study of catastrophic forgetting to the scenarios described in section \ref{ssec:CLsce}, where every task represents a supervised classification problem, the proposed method can be applied to many other settings. For any continual learning task, which involves training an ANN to minimize a loss functions over a sequence of tasks, the change in loss can be approximated as described above. Examples for such sequences of task include, but are not limited to, learning of representations over different domains, a sequence of regression tasks or training generative models continually to capture different data generating distributions.

\section{Experiments}
\label{sec:exper}

In this section we will introduce the model used for the following experiments and describe the exact realization of the different continual learning scenarios introduced in section \ref{ssec:CLsce}. We use the same architecture on all three scenarios in order to allow for a comparison of the results. Since a ANNs architecture can have a significant influence on its resilience to catastrophic forgetting \cite{goodfellow2013empirical} and we are interested in analyzing what challenges the different scenarios pose to an ANN, changing the models structure between evaluations is avoided. The architecture used in this work is a small convolutional neural network (CNN) with four hidden layers according to table \ref{tbl:CNN}. Dropout is applied to the input of respective layers and a stride of $2$ is used in all convolutions.

\begin{table}
	\caption{Model architecture}
	\vspace{0.3cm}
	\begin{tabular}{cccc}
		\hline 
		\multicolumn{4}{c}{CNN Architecture} \\ 
		\hline 
		Layer & Act. Size & Act. Func. & Dropout \\ 
		\hline 
		Input & $28\times 28\times 1$ & - & - \\ 
		Conv. $32\times 3\times 3$& $14\times 14\times 32$ & ReLU & - \\ 
		Conv. $32\times 3\times 3$& $7\times 7\times 32$ & ReLU & - \\ 
		Dense & $64$ & ReLU & $0.2$ \\ 
		Dense & $32$ & ReLU & $0.2$ \\ 
		Dense & $10$ & Soft max & $0.2$ \\ 
		\hline
	\end{tabular}
	\label{tbl:CNN}
\end{table} 

\subsection{Incremental Task Learning}
\label{ssec:IncTa}

As described in section \ref{ssec:CLsce}, ITL is characterized by a change in the domain $\mathcal{D}$ and the assignment $\mathcal{A}$. In order to generate two tasks, which can be used to localize catastrophic forgetting in this scenario, we utilize two popular data sets, MNIST \cite{lecun-mnisthandwrittendigit-2010} and FashionMNIST \cite{xiao2017/online}. While MNIST is a data set for handwritten digit classification with $60000$ training and $10000$ test samples of size $28\times 28\times 1$, FashionMNIST is a drop in replacement for MNIST containing images of fashion from 10 different categories.\\\\
The sequence of tasks is created by first training the ANN on classifying the handwritten digits of MNIST and then the different fashion categories of FashionMNIST. During this sequence the domain and assignment change. Considering the domains of both tasks, $\mathcal{D}_{M}$ and $\mathcal{D}_{F}$, the feature space $\mathcal{X}_{M}=\mathcal{X}_{F}=\lbrace 0,1,\ldots,255\rbrace^{28\times 28\times 1}$ remains unchanged, while the data generating distributions change, i.e. $\mathcal{P}_{M}(\mathbf{X})\neq\mathcal{P}_{F}(\mathbf{X})$. The assignments, $\mathcal{A}_{M}$ and $\mathcal{A}_{F}$, differ in both the label space and the function learned from training examples, i.e. $\mathcal{Y}_{M}\neq\mathcal{Y}_{F}$ and $f_{M}\neq f_{F}$. We realize this by utilizing a separate output layer, with $10$ neurons, of the ANN for each task.

\subsection{Incremental Domain Learning}
\label{ssec:IncDo}

In IDL the domain $\mathcal{D}$ changes between tasks while the assignment $\mathcal{A}$ is unchanged. This means, although the feature space $\mathcal{X}$ and/or the corresponding data generating distribution $P(\mathbf{X})$ change, the ANN has to solve the same task but based on different inputs. For experiments on IDL we again utilize the MNIST data set.\\\\
A common way to generate a sequence of tasks for IDL based on the MNIST data set is to apply a different random permutation of pixels to each image in the data set for every new task. This is known as permutation MNIST in the literature \cite{kirkpatrick2017overcoming, zenke2017continual, shin2017continual, v.2018variational, Ven2018GenerativeRW}. Formally we change the domain $\mathcal{D}$ between tasks by changing the data generating distribution $P(\mathbf{X})$, while the feature space $\mathcal{X}=\lbrace 0,1,\ldots,255\rbrace^{28\times 28\times 1}$ remains unchanged. It is possible to create a very large number of different tasks using this method, but most of the generated domains are uncorrelated and do not resemble natural images. 

\subsection{Incremental Class Learning}
\label{ssec:IncCl}

Experiments for ICL are commonly based on learning the classes in one specific data set in an incremental way. This typically corresponds to a change in both the domain $\mathcal{D}$ and the assignment $\mathcal{A}$ between tasks. But, in contrast to ITL, the feature space $\mathcal{X}$ and the label space $\mathcal{Y}$ are shared between the tasks.\\\\
In order to generate a sequence of tasks for ICL, a data set is commonly split into disjoint subsets, where each subset contains only examples of one or more classes. A widely used example for this is split MNIST \cite{zenke2017continual, shin2017continual, v.2018variational, Ven2018GenerativeRW}. In this case the MNIST data set is commonly split into 5 subsets containing two classes each, e.g. $(0,1)$, $(2,3)$, $(4,5)$, $(6,7)$ and $(8,9)$. Learning to classify the classes in each of these subsets is considered a task.

\section{Results}
\label{sec:resul}

\begin{figure}
	\centering
	\includegraphics[width=1.0\linewidth]{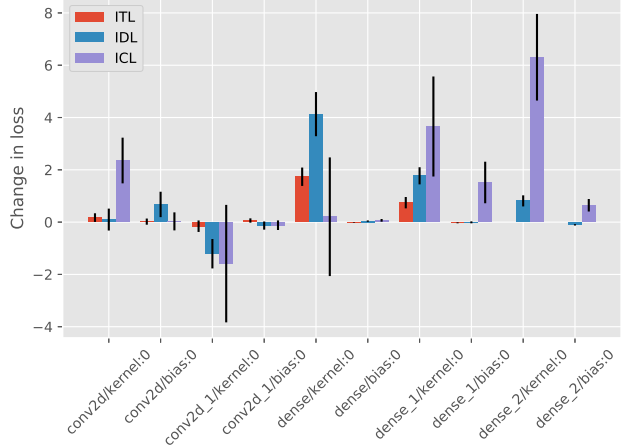}
	\caption[Results of experiments]{Results of the experiments described in section \ref{sec:exper}. This plot shows the absolute contribution of the weight matrices and bias vectors of individual layers. The ordering from left to right mirrors the CNN's structure from table \ref{tbl:CNN}.}
	\label{fig:combinedresultssum}
\end{figure}

\begin{figure}
	\centering
	\includegraphics[width=1.0\linewidth]{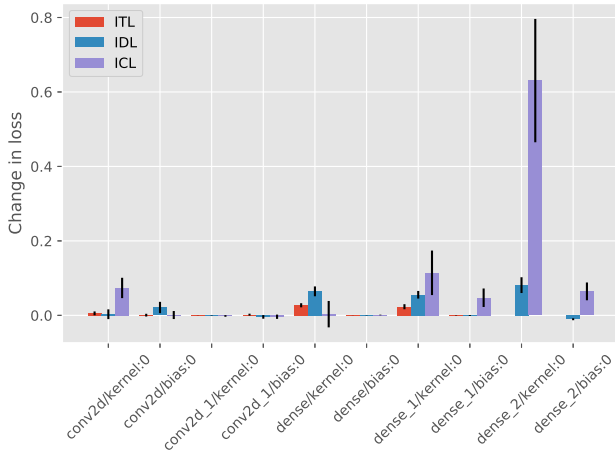}
	\caption[Results of experiments]{Results of the experiments described in section \ref{sec:exper}. This plot shows the average contribution of an element in the weight matrices and bias vectors of individual layers. The ordering from left to right mirrors the CNN's structure from table \ref{tbl:CNN}.}
	\label{fig:combinedresultsmean}
\end{figure}

In this section we present the results obtained during the experiments described in section \ref{sec:exper}. We use the CNN architecture depicted in table \ref{tbl:CNN} and train with the Adam optimizer and a learning rate of $0.001$ and a batch size of $128$ for $10$ epochs on every task. No learning rate schedules or early stopping was used. Since the extend of catastrophic forgetting depends not only on the architecture used but also the random initialization of the model we run every experiment $10$ times and report average values with their standard deviation. Although an interpretation of the results is difficult, since it is highly dependent on many different factors like the model architecture, weight initialization, the optimizer and many other hyperparameters, we can at least compare the same configuration across the three different continual learning scenarios introduced in section \ref{ssec:CLsce}.\\\\
Figure \ref{fig:combinedresultssum} shows the absolute change in loss aggregated over the weight matrices/tensors and bias vectors of every layer. The ordering from left to right corresponds with the architecture shown in table \ref{tbl:CNN} from top to bottom. Comparing the overall distribution of change in loss over the different continual learning scenarios, we can observe distinct patterns for these. The absolute contributions of the convolutional layers on all scenarios is lower than those of the dense layers. Also the absolute contribution of the bias vectors is small when compared to the weight matrices. We can even identify an average decrease in loss for the weight tensor of the second convolutional layer. On ICL the variance of this change in loss is however very high when compared to the other scenarios. While the convolutional layers show a more or less homogeneous change over the different continual learning scenarios, we can observe an interesting difference in the dense layers across them. On ITL and IDL the absolute contribution of the layers decreases from left to right. This is expected since the overall number of neurons in these layers also decreases from left to right. On ICL however we can observe an opposite behavior. Although the number of neurons decreases, the over all contribution to the change in loss increases. This observation is in line with Farguhar \& Gal \cite{Farquhar2018TowardsRE} who observe more catastrophic forgetting on split MNIST than on permutation MNIST and reason that this is caused by gradients with higher magnitude while training the last layer due to more similar looking images in ICL when compared with IDL. Although our observations also indicate that in ICL the last layers are mostly responsible for the change in loss and therefore catastrophic forgetting, we can not give a general explanation for this when considering the limited scope of our experiments. But we can at least support Farguhar \& Gal observation that the last layers are mostly responsible for catastrophic forgetting in ICL. This becomes even more evident when we consider the average contribution of a neuron/filter over the layers as shown in figure \ref{fig:combinedresultsmean}. Here we have averaged the contributions of neurons, filters without their bias elements, which are plotted separately. Comparing ITL, IDL and ICL we can again observe that the average contribution of a neuron/filter increases when going from ITL to IDL and reaches its maximum for ICL. But we can also again observe that while on ITL and IDL the average contribution of a neuron/filter is approx constant over the dense layers while on ICL it increases from the first dense layer to the output layer.\\\\
Overall we can observe different responses of the studied architecture when exposed to the three continual learning scenarios. While ITL causes the least catastrophic forgetting, the evaluation of IDL shows very similar behavior but increased catastrophic forgetting. Our evaluation on ICL not only shows the overall highest change in loss but also a very different pattern than the other two scenarios.

\section{Conclusion}
\label{sec:concl}

Catastrophic forgetting is a fundamental problem in the training process of ANNs. Although it was studied in the past, there was surprisingly few research on the phenomenon itself published over the recent years. We proposed a method for determining the contribution of individual parameters in an ANN to a change in loss, which can be linked to catastrophic forgetting. This method allows a more detailed analysis of this phenomenon through a localization of parts in an ANN that contribute the most to such a change in loss. We evaluated our method on three different continual learning scenarios on common data sets in the field. We could not only support claims from other researchers based on a different experimental evaluation but also found similarities and differences in the response of a specific ANN, which was exposed to these different scenarios. 

\vfill\pagebreak

\bibliographystyle{IEEEbib}
\bibliography{refs}

\end{document}